\documentclass[10pt]{article}
\usepackage[margin=1in]{geometry}

\usepackage{amsmath}
\usepackage{bm}
\usepackage{amssymb}
\usepackage[linesnumbered,ruled,vlined,nosemicolon]{algorithm2e}
\usepackage[numbers]{natbib}
\usepackage{hyperref}
\usepackage{mathtools}
\usepackage{appendix}
\usepackage{tabularx}
\usepackage{xcolor}
\usepackage{pifont}

\SetKwInput{KwInput}{Input}
\SetKwInput{KwOutput}{Output}

\title{Power side-channel leakage localization through adversarial training of deep neural networks}
\author{Jimmy Gammell\thanks{Corresponding author} \quad Anand Raghunathan \quad Kaushik Roy\\Elmore Family School of Electrical and Computer Engineering\\Purdue University\\West Lafayette, IN, United States\\\texttt{\{jgammell, araghu, kaushik\}@purdue.edu}}
\date{}

\DeclareMathOperator{\sbytes}{SBOX}

\DeclareMathOperator{\avg}{Average}
\DeclareMathOperator{\std}{StandardDeviation}
\DeclareMathOperator{\zscore}{Z-score}

\DeclarePairedDelimiter{\set}{\{}{\}}

\DeclarePairedDelimiter{\abs}{\lvert}{\rvert}
\DeclarePairedDelimiter{\seq}{\{}{\}}

\newcommand{\xor}{\oplus}
\newcommand{\zint}[2]{\left\{#1, \hdots, #2\right\}}
\newcommand{\defeq}{:=}

\newcommand{\setfnt}{\mathsf}

\newcommand{\refsec}[1]{Sec. \ref{#1}}
\newcommand{\reffig}[1]{Fig. \ref{#1}}

\newcommand{\refalg}[1]{Alg. \ref{#1}}
\newcommand{\refalgln}[1]{Ln. \ref{#1}}
\newcommand{\reftbl}[1]{Tbl. \ref{#1}}

\begin{document}

\maketitle

\begin{abstract}
    Supervised deep learning has emerged as an effective tool for carrying out power side-channel attacks on cryptographic implementations. While increasingly-powerful deep learning-based attacks are regularly published, comparatively-little work has gone into using deep learning to defend against these attacks. In this work we propose a technique for identifying which timesteps in a power trace are responsible for leaking a cryptographic key, through an adversarial game between a deep learning-based side-channel attacker which seeks to classify a sensitive variable from the power traces recorded during encryption, and a trainable noise generator which seeks to thwart this attack by introducing a \textit{minimal} amount of noise into the power traces. We demonstrate on synthetic datasets that our method can outperform existing techniques in the presence of common countermeasures such as Boolean masking and trace desynchronization. Results on real datasets are weak because the technique is highly sensitive to hyperparameters and early-stop point, and we lack a holdout dataset with ground truth knowledge of leaking points for model selection. Nonetheless, we believe our work represents an important first step towards deep side-channel leakage localization without relying on strong assumptions about the implementation or the nature of its leakage. A PyTorch implementation of our algorithm and experiments can be found \href{https://github.com/jimgammell/gan_side_channel_leakage_detector/tree/main}{here}.
\end{abstract}

\section{Introduction}

In recent years, supervised deep learning \cite{lecun2015} has emerged as a powerful strategy for carrying out side-channel attacks (SCAs) \cite{EPRINT:MagPorPro16}. Deep neural network (DNN)-based SCAs regularly achieve comparable or superior performance compared to traditional attacks, despite requiring significantly less manual data preprocessing based on \textit{a priori} knowledge about the attacked cryptographic device \cite{EPRINT:MagPorPro16, EPRINT:CagDumPro17, TCHES:LZCGL21}. These attacks pose a major threat to implementations of cryptographic algorithms such as the ubiquitous advanced encryption standard (AES) \cite{daemen1998} and other symmetric-key algorithms whose secret keys can be determined by observing power consumption or electromagnetic emissions over time while plaintexts are encrypted.

Despite this threat, DNN-based SCAs have a unique potential to elucidate the vulnerabilities in cryptographic implementations and enable the design of targeted countermeasures to mitigate them. While traditional techniques such as template attacks \cite{CHES:ChaRaoRoh02} and differential power analysis \cite{C:KocJafJun99} can provide a lower bound on the extent to which a device is vulnerable to power SCAs and may identify a subset of the timesteps responsible for this vulnerability, they are limited by their reliance on manual preprocessing, such as extraction of a small number of `points of interest' from raw power traces, knowledge of random Boolean mask values at profiling-time, or synchronization of traces. Because of this, these methods may not necessarily exploit sources of leakage which were not accounted for during the preprocessing stage. In contrast, supervised DNN-based SCAs are effective even when little or no preprocessing is done, enabling the learned classifiers to utilize more sources of leakage. It is then possible to analyze the resulting learned classifiers to identify sources of leakage which were unknown prior to the attack.

In this work we propose a strategy to localize power trace side-channel leakage through an adversarial game played between a DNN-based classifier and a trainable noise generator, in a manner similar to generative adversarial networks \cite{goodfellow2014}. The classifier learns to predict the value of some sensitive variable using power traces, while the noise generator learns to introduce noise into the power traces to optimize a compromise between degrading classifier performance and adding as little noise as possible, in a sense we will subsequently precisely define. At the end of training, lots of noise will be added to the timesteps which have high utility for predicting the sensitive variable and little noise will be added to those with low utility, and on this basis the leaking timesteps can be identified. The classifier can take advantage of state-of-the-art DNN-based SCA techniques, allowing the efficacy of our leakage localization strategy to scale with the ever-increasing efficacy of attack strategies as they are published.

To our knowledge, most existing leakage localization strategies rely either on first-order statistics of the power traces, or on `neural network attribution' techniques applied to DNNs which have been trained to carry out an SCA. In contrast to the former, our technique can identify a dependency between a sensitive variable and multiple timesteps of the input power trace, even when the sensitive variable is pairwise-independent of these timesteps and there is no \textit{a priori} knowledge that the dependency exists (e.g. when Boolean masking is used to protect an attack point). It can also be more data-efficient when useful inductive biases can be imposed through choice of the classifier architecture and the objective function (e.g. convolutional architectures with pooling can bias the classifier towards temporal-translation-invariant predictors to mitigate the effects of trace desynchronization, and an L1-norm penalty on the output of the noise generator can bias it towards producing temporally-sparse noise). In contrast to the latter, our technique has a reduced tendency to falsely identify timesteps as non-leaking, because whereas there is no guarantee that a successful DNN-based attacker will use every dependency between the sensitive variable and power trace timesteps in its predictions, in our technique the classifier is trained simultaneously with an adversarial noise generator, and is forced to seek out new dependencies as those it presently relies on are be progressively attenuated by noise.

In summary, the contribution of this work is:
\begin{itemize}
    \item We propose a technique for localizing the power trace timesteps responsible for leaking the cryptographic key to power side-channel attacks on AES implementations, through an adversarial game played between a deep neural network (DNN)-based side-channel attacker and a trainable noise generator seeking to thwart the attack. To our knowledge, this is the first proposed leakage localization technique based on an adversarial game played with a DNN.
    \item We demonstrate experimentally that in contrast to first-order statistics-based leakage localization techniques, our technique can localize leakage of Boolean-masked sensitive variables without profiling-time knowledge of the mask values. It also has improved data-efficiency when applied to desynchronized power traces.
    \item We demonstrate experimentally that compared to neural network attribution-based leakage localization techniques, our technique has a reduced tendency to fail to recognize sources of leakage when many are present.
\end{itemize}

We will subsequently present the methodology for our technique and for representative examples of leakage localization techniques based on first-order statistical techniques and neural network attribution. We will then experimentally show that our technique outperforms these baselines on synthetic datasets when common countermeasures are present. A major limitation of the present method is its high sensitivity to hyperparameters and early-stopping point, and it is thus challenging to use it on real datasets where we lack a holdout dataset with ground truth knowledge of leaking points for model selection. Nonetheless, we present results on the publicly-available DPAv4 and ASCADv1-fixed datasets with model selection based on visual inspection, and attain qualitatively-reasonable results.

\section{Background and related work}

\subsection{Supervised deep learning for profiled power side-channel attacks}

\subsubsection{Profiled power side-channel attacks}

Symmetric-key cryptographic algorithms such as AES \cite{daemen1998} take as input a key $k'\in\mathbb{Z}_+$ and plaintext $p'\in\mathbb{Z}_+$ and return a ciphertext $c'\in\mathbb{Z}_+$, where $p'$ can be uniquely determined from $c'$ given $k'$ but is statistically-independent of $c'$ when nothing is known about $k'$. This allows parties with exclusive knowledge of $k'$ to communicate over insecure channels without fear of eavesdroppers learning the communicated information, by encrypting the communications according to $k'$, transmitting the resulting ciphertexts, and decrypting the ciphertexts upon receipt. In practice, the security of such communications hinges on $k'$ remaining a secret value known only by the senders and intended recipients of encrypted messages.

While AES and other cryptographic algorithms are secure in the sense that there is no known computationally-feasible algorithm to determine $k'$ by encrypting plaintexts and observing the resulting ciphertexts, \textit{physical implementations} of cryptographic algorithms inevitably leak information through measurable physical signals which are statistically-dependent on the key, called \textit{side-channels}. Side-channels include quantities such as instantaneous power consumption during encryption, electromagnetic radiation during encryption, and algorithm execution time. There is a large body of work on side-channel attacks \cite{C:Kocher96, C:KocJafJun99, CHES:ChaRaoRoh02, CHES:BriClaOli04}, which are algorithms to determine $k'$ by observing side-channel emissions during encryptions. In this work we consider \textit{profiled power side-channel attacks}, where the attacker has access to a clone of the target cryptographic implementation and repeatedly measures the instantaneous power consumption during encryption after specifying a plaintext and key, in order to model the probabilistic relationship between power consumption and the key.

When attacking AES implementations, instead of predicting $k'$ directly it is common to learn a predictor of an `attack point', which is an intermediate variable of the cryptographic algorithm which provides information about $k'$ and strongly influences the power consumption due to being directly-manipulated. A common attack point choice is $z \defeq \sbytes(k \xor p)$, where $k\in\zint{0}{255}$ and $p\in\zint{0}{255}$ denote individual bytes of $k'$ and $p'$ and $\sbytes$ is an invertible function which is publicly-available. Given knowledge of $p$, $z$ uniquely determines the corresponding byte of $k'$ through the identity $k = \sbytes^{-1}(z) \xor p$. The attack can be repeated for each pair of bytes $(p, k)$ of $p'$ and $k'$ to recover the full key. In this work we will consider leakage of one such attack point $z$.

\subsubsection{Profiling through supervised learning}

Given a clone of the target device, an attacker can choose plaintexts and keys uniformly at random and measure the power traces while running the corresponding encryptions, to compile a dataset
\begin{equation}
    \setfnt{D} \defeq \{(x_i, z_i)\}_{i=1}^{N} \in \left(\mathbb{R}^d \times \zint{0}{255}\right)^N
\end{equation}
where the values $x_i\in\mathbb{R}^d$ denote measured power traces and $z_i\in\zint{0}{255}$ denote the aforementioned attack points for one pair of key and plaintext bytes. The attacker can then assume the existence of some underlying data-generating distribution $p(x, z)$ on $\mathbb{R}^d\times\zint{0}{255}$, of which $\setfnt{D}$ consists of i.i.d. samples.

In such a scenario, there is a wide body of work on \textit{supervised learning} techniques \cite{bishop2006} for using $\setfnt{D}$ to find a predictor of attack points given their associated plaintexts, e.g. a function $h:\mathbb{R}^d\to\Delta\zint{0}{255}$ where $\Delta \zint{0}{255}$ denotes the set of probability mass functions on $\zint{0}{255}$, for which $h(x) \approx p(\cdot | x)$ with high probability for $x$ generated by the data distribution $p$. These techniques entail defining a hypothesis space $\setfnt{H}\subset\mathbb{R}^d\to\Delta\zint{0}{255}$ of candidate functions, then searching $\setfnt{H}$ for a function which minimizes an objective function constructed to be minimal for `high-quality' functions, according to some desired definition of `quality'. A common choice is to minimize the \textit{empirical risk minimization} objective,
\begin{equation}
    h^*\in\arg\min_{h\in\setfnt{H}} \; \mathcal{L}(h) \quad \text{where} \quad \mathcal{L}(h) \defeq \frac{1}{N}\sum_{i=1}^{N} \ell(h(x_i), z_i),
\end{equation}
where $\ell:\Delta\zint{0}{255}\times\zint{0}{255}\to\mathbb{R}_+$ is a loss function which returns smaller values when $h(x_i)$ assigns higher mass to $z_i$.

\subsubsection{Supervised deep learning}

Over the past decade, deep learning-based techniques \cite{lecun2015} have underpinned major advances in diverse domains including natural language processing \cite{ouyang2022}, photo-realistic image synthesis \cite{brock2018, dhariwal2021}, computer vision \cite{krizhevsky2012, dosovitskiy2020}, playing video games \cite{mnih2015} and board games \cite{silver2017}, and protein structure prediction \cite{jumper2021}, despite requiring little expert human knowledge or manual data preprocessing compared to prior approaches. Interest has consequently surged \cite{EPRINT:PPMWB21} in deep learning-based side-channel analysis, with recent work exploring topics such as finding efficient and performant neural network architectures for this domain \cite{EPRINT:MagPorPro16, JCEng:BPSCD20, EPRINT:ZBHV19, TCHES:WAGP20}, learning from protected AES implementations with little or no manual preprocessing to circumvent countermeasures \cite{JCEng:BPSCD20, EPRINT:CagDumPro17, masure2023}, and learning from raw traces without point-of-interest selection \cite{TCHES:LZCGL21, bursztein2023}.

A deep neural network (DNN) is a parametric function approximator $\Phi:\mathbb{R}^d\times\mathbb{R}^p\to\mathbb{R}^{256},\;\;(x, \theta) \mapsto \Phi(x; \theta)$ composed by interleaving many parametric linear and nonlinear functions, which can approximate any continuous function from a compact subset of $\mathbb{R}^d$ to $\mathbb{R}^{256}$ to within arbitrarily-small error, provided it has an appropriate architecture. Details about common neural network architectures can be found in textbooks such as \cite{murphy2023}.

When using DNNs for supervised learning tasks, we implicitly define the hypothesis space in terms of the neural network architecture, $\setfnt{H} = \{\Phi(\cdot; \theta) : \theta\in\mathbb{R}^p\}$, and redefine the objective, which is typically the empirical risk, in terms of $\theta$:
\begin{equation}
    \mathcal{L}(\theta) \defeq \frac{1}{N} \sum_{i=1}^{N} \ell(\Phi(x_i; \theta), z_i).
\end{equation}
We typically choose $\ell$ to be differentiable and minimize $\mathcal{L}$ using \textit{minibatch stochastic gradient descent (SGD)}, where we randomly partition $\zint{1}{N}$ into minibatch index sets $\setfnt{B}_1, \dots, \setfnt{B}_{T}$ and then iteratively compute the parameters
\begin{equation}
    \theta_{t+1} = \theta_t - \eta \tilde{\nabla}\mathcal{L}(\theta_t) \quad \text{where} \quad \tilde{\nabla}\mathcal{L}(\theta_t) = \frac{1}{|\setfnt{B}_t|}\sum_{i\in\setfnt{B}_t} \left[\nabla_\theta \ell(\Phi(x_i; \theta), z_i)\right]_{\theta=\theta_t}
\end{equation}
and $\eta>0$ is a scalar called the \textit{learning rate}.

In practice this is usually repeated for multiple passes over $\setfnt{D}$, called epochs. It is common to use variants of the minibatch SGD algorithm incorporating momentum and adaptive per-parameter learning rates (in this work we use the Adam optimizer \cite{kingma2015}). Since there are generally a vast number of parameters $\theta\in\mathbb{R}^p$ which perfectly minimize the empirical risk but achieve varying levels of performance in expectation over data generated by $p$, it is common to modify $\mathcal{L}$ and the training algorithm, as well as the architecture $\Phi$, to bias the iterates $\theta_t$ towards regions of $\mathbb{R}^p$ in which $\Phi(\cdot;\theta_t)$ is expected \textit{a priori} to generalize well, or is observed to generalize well on an i.i.d. validation dataset which is not used in the computation of $\mathcal{L}$ or its gradient estimates.

\subsection{Power side-channel leakage localization}

Existing leakage localization approaches can broadly be categorized into those based on first-order statistical techniques, and those based on neural network attriburion techniques.

\subsubsection{First-order statistical techniques}
\label{sec:snr}
First-order statistical techniques are often used as an initial `point-of-interest' detection step to reduce the dimensionality of raw power trace measurements prior to applying SCA techniques. Given a database of power traces in $\mathbb{R}^d$, the goal is to identify timesteps $\tau_1, \dots, \tau_m \in \zint{1}{d}$ at which power consumption has a strong dependence on the value of some sensitive variable, where $m<\!\!<d$. The motivation is normally to reduce the computational cost of modeling the relationship between a power trace and the cryptographic key, given long power traces in which most measurements are nearly-independent of the key.

The general approach is to compute the average power measurement at each timestep given various distinct values of the sensitive variable (e.g. a byte, byte Hamming weight, or bit of the first SBOX operation of the AES algorithm), then assign a score to each timestep which is larger for timesteps at which the power measurement varies more for different sensitive variable values. In this work we consider the signal-noise ratio (SNR) \cite{mangard2008}, defined in \refalg{alg:snr_computation}, as a representative example of the category of first-order statistical leakage localization techniques, but there are many other examples, such as the sum of differences \cite{CHES:GieLemPaa06}, t-statistic \cite{welch1947, mather2013} and the $\chi^2$ test \cite{TCHES:MRSS18}.

\begin{algorithm}[h!]
    \KwInput{$\setfnt{D}=\seq{(x^{(i)}, z^{(i)})}_{i=1}^{N}\in \left( \mathbb{R}^d\times\zint{0}{255} \right)^N$ \tcp{dataset of traces + attack point pairs}}
    \KwOutput{$m\in\mathbb{R}^d$ \tcp{signal-noise ratio mask}}
    \BlankLine
    \For{$z\in\zint{0}{255}$}{
        Compute $\mu_z \gets \avg_{i\in\zint{1}{N} : z^{(i)}=z} x^{(i)}$ \tcp{average power trace for fixed attack point value}
    }
    Compute $\mu \gets \avg_{i\in\zint{1}{N}} x^{(i)}$ \tcp{average power trace over full dataset}
    Compute $\sigma^2_{\text{signal}} \gets \avg_{z\in\zint{0}{255}} (\mu_z - \mu)^2$ \tcp{signal variance}
    Compute $\sigma^2_{\text{noise}} \gets \avg_{i\in\zint{1}{N}} (x^{(i)} - \mu_{z^{(i)}})^2$ \tcp{noise variance}
    Compute $m \gets \sigma^2_{\text{signal}} / \sigma^2_{\text{noise}}$ \tcp{signal-noise ratio mask}
    \Return $m$
\caption{Signal-to-noise ratio (SNR) computation}
\label{alg:snr_computation}
\end{algorithm}

The main shortcoming of these techniques is that because they consider individual timesteps in isolation, they fail to detect dependencies between \textit{multiple} timesteps and the cryptographic key. For example, \textit{Boolean masking} is a common countermeasure against side-channel attacks, where instead of operating directly on a sensitive byte $z\in\zint{0}{255}$, the algorithm randomly draws a `Boolean mask' byte $r\sim\mathcal{U}\zint{0}{255}$ and then operates on $z \xor r$ where $\cdot \xor \cdot$ denotes the bit-wise XOR operation. In this context, $z$ is independent of $z \xor r$ absent knowledge of $r$, but can be determined with certainty given $r$. Because $r$ and $z \xor r$ influence different timesteps in the power trace, techniques considering the timesteps in isolation will fail to detect this source of leakage. Many works sidestep this limitation by assuming knowledge of the Boolean mask values at profiling-time, but since these mask values are intermediate variables which are likely to be hidden from users of a cryptographic implementation, in a practical attack such knowledge may be unattainable or attainable only through great effort.

\subsubsection{Neural network attribution techniques}
\label{sec:gradvis}

Another approach for leakage localization is to train a DNN-based side-channel attacker using standard supervised learning techniques, then use one of a variety of `neural network attribution' techniques to score the timesteps of the input power traces in terms of their influence on the classification decision. Given a database of power traces $x_1, \dots, x_N \in \mathbb{R}^d$ and a trained classifier $\Phi^*:\mathbb{R}^d\to\mathbb{R}^{256}$, these techniques generate importance masks for each power trace, $m_1, \dots, m_N \in \mathbb{R}^d$, where the score of a timestep increases with its amount of `influence' on the predictions made by $\Phi^*$, which can then be summarized by averaging over the dataset.

Many methods exist for generating these importance masks given a pretrained classifier $\Phi^*$, including differentiating functions of the output of $\Phi^*$ with respect to its input \cite{EPRINT:MasDumPro18, SAC:HetGehGun19}, occluding sections of the input \cite{SAC:HetGehGun19}, ablating sections of neural networks \cite{EPRINT:WWJPBP21}, and singular vector canonical correlation analysis of internal representations \cite{valk2021}. In this work, we consider the gradient visualization technique (GradVis) \cite{EPRINT:MasDumPro18}, defined in \refalg{alg:gradvis_computation}, as a representative example.

\begin{algorithm}[h!]
    \KwInput{$\setfnt{D}=\seq{(x^{(i)}, z^{(i)})}_{i=1}^{N}\in\left(\mathbb{R}^d\times\zint{0}{255}\right)^N$ \tcp{dataset of traces + attack point values}}
    \KwInput{$\Phi^*:\mathbb{R}^d\to\mathbb{R}^{256}$ \tcp{deep neural network trained on $\setfnt{D}$ using standard supervised learning techniques}}
    \KwInput{$\ell:\Delta\zint{0}{255}\times\zint{0}{255}\to\mathbb{R}_+$ \tcp{loss function used to define the empirical risk when the DNN was trained}}
    \KwOutput{$m\in\mathbb{R}^d$ \tcp{Gradient visualization mask}}
    \BlankLine
    \For{$i\in\set{1, \dots, N}$}{
        Compute $\tilde{m}^{(i)} \gets \left.\nabla_x \ell\left(\Phi^*(x), z^{(i)}\right)\right\rvert_{x=x^{(i)}}$ \tcp{sensitivity of loss to small perturbation of each input element}
    }
    Compute $m \gets \avg_{i\in\zint{1}{N}} \abs{\tilde{m}^{(i)}}$ \tcp{average sensitivity over full dataset}
    \Return $m$
\caption{Gradient visualization (GradVis) computation}
\label{alg:gradvis_computation}
\end{algorithm}

While these techniques avoid the main shortcomings of first-order statistical techniques, they have the major shortcoming that there is no guarantee that $\Phi^*$ will rely on \textit{all} relevant timesteps when classifying power traces, meaning that the resulting importance mask will assign high importance to only a subset of the timesteps which may be used to identify the cryptographic key. It has been documented in other domains that DNNs often ignore input-output relationships which may be useful for prediction; for example, CNN-based image classifiers are biased towards textures and often fail to capture relationships between object shapes and labels \cite{geirhos2018}, DNN classifiers have been shown to ignore XOR relationships when linear relationships are present \cite{hermann2020}, and more-generally to ignore complex input-output relationships when simpler relationships (e.g. more linearly-decodable from the hidden DNN activations at initialization) are present \cite{geirhos2020}.

\subsection{Other related work}

Our approach of joint adversarial training of a DNN-based classifier and a noise generator was inspired by generative adversarial networks (GANs) \cite{goodfellow2014}. While we are not aware of existing work applying GAN-like approaches to leakage localization, GANs have been used for data augmentation to facilitate power side-channel attacks \cite{wang2020, mukhtar2022}.

Leakage localization is similar to weakly-supervised semantic segmentation from labels \cite{ahn2018}, which is a well-studied problem in computer vision. In this setting, the goal is to learn a model which highlights pixels corresponding to a particular class in images, by training on image data with class-level labels but no pixel-level labels. Many techniques in this domain are based on attribution of DNN classifier predictions while adversarially-erasing regions of input images to encourage the classifier to use all discriminative regions of inputs \cite{wei2017, zhang2018, kweon2021}. These techniques are conceptually-similar to ours, the major difference being that they use non-persistent per-image adverarial masks whereas we use a single adversarial mask for the entire dataset which is incrementally trained.

GANs have been applied to `data sanitization', where a `generator' DNN seeks to minimally modify inputs while removing a dependence between the inputs and some sensitive variable. Similarly to our technique, this is achieved through an adversarial game played against a classifier DNN which tries to predict the sensitive variable from the inputs. This approach has been used to promote fairness by sanitizing datapoints or DNN activations of dependence on a protected attribute such as sex or race \cite{edwards2015, hwang2020, zhang2018, xie2017, madras2018}, and to enable robustness to data distribution shifts by sanitizing inputs or DNN activations of dependence on properties of data which may change when the data distribution changes \cite{li2018, li2018_a, deng2020}. Our technique differs from these because instead of a DNN-based `generator' we have a data-independent mask, and we seek a mask which is applicable to the full dataset rather than specialized for individual datapoints within it.

\section{Methodology}

\subsection{Adversarial masking (AdvMask)}

\begin{figure}
    \centering
    \includegraphics[width=\textwidth]{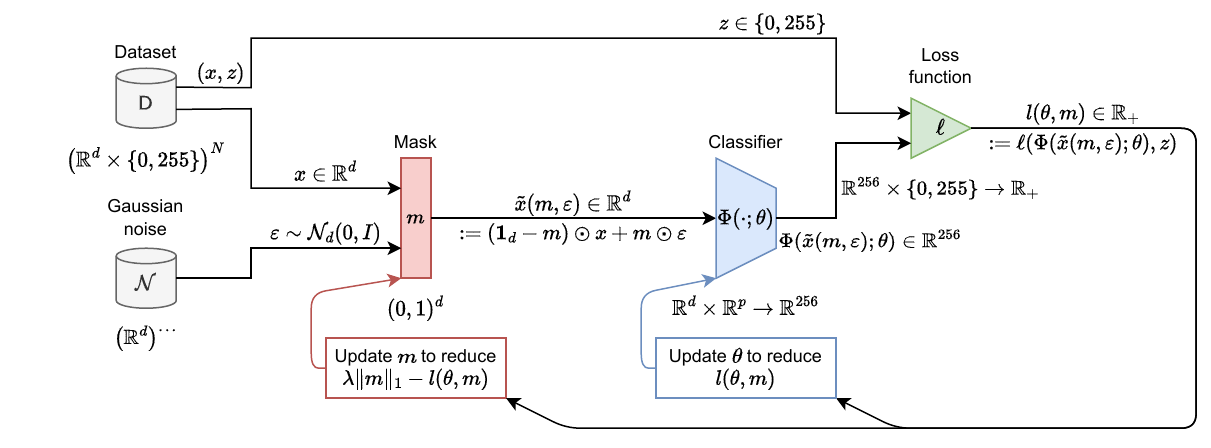}
    \caption{Schematic illustration of AdvMask, our adversarial masking technique for side-channel leakage localization.}
    \label{fig:advmask}
\end{figure}

In \refalg{alg:advmask} (illustrated schematically in \reffig{fig:advmask}) we present our AdvMask technique for side-channel leakage localization through simultaneous adversarial training of a DNN-based classifier and a noise generator. The classifier is trained in a supervised manner to classify noisy power traces based on the value of a sensitive variable, which in this work is the output of the first SBOX operation of the AES algorithm. Simultaneously, a noise generator, represented by a data-independent vector $m\in (0, 1)^d$ which dictates a convex combination between power trace timesteps $x\in\mathbb{R}^d$ and i.i.d. $d$-dimensional standard normal noise $\varepsilon\sim\mathcal{N}_d(0, I)$, is trained to optimize a composite objective function which rewards both increasing the loss of the classifier and penalizing the L1 norm of $m$ $\lVert m \rVert_1 \defeq \sum_{i=1}^{d} \lvert m_i \rvert$. We optimize these objectives using alternating minibatch stochastic gradient descent, which is similar to the training procedure for generative adversarial networks \cite{goodfellow2014} and can be easily implemented using the popular deep learning libraries such as PyTorch or Keras.

\begin{algorithm}[h!]
    \SetKwFunction{ApplyMask}{ApplyMask}
    \KwInput{$\setfnt{D}=\seq{(x^{(i)}, z^{(i)})}\in\left(\mathbb{R}^d\times\zint{0}{255}\right)^N$ \tcp{dataset of traces + attack point values}}
    \KwInput{$\Phi:\mathbb{R}^d\times\mathbb{R}^p\to\mathbb{R}^{256},\;\;(x,\theta)\mapsto\Phi(x; \theta)$ \tcp{classifier neural network architecture}}
    \KwInput{$\ell:\mathbb{R}^{255}\times\zint{0}{255}\to\mathbb{R}_+$ \tcp{loss function used to define empirical risk of DNN classifier (default: categorical cross-entropy loss)}}
    \KwInput{$\operatorname{UpdateParams}^\theta:\mathbb{R}^p\to\mathbb{R}^p$ \tcp{parameter update function for the classifier (default: Adam optimizer with $\alpha=10^{-3}, \beta_1=0.9, \beta_2=0.999$)}}
    \KwInput{$\operatorname{UpdateParams}^{m}:\mathbb{R}^d\to\mathbb{R}^d$ \tcp{parameter update function for the mask (default: Adam optimizer with $\alpha=10^{-2}, \beta_1=0.5, \beta_2=0.0$)}}
    \KwInput{$\theta_0 \in \mathbb{R}^p$ \tcp{initial classifier parameters (default: $\theta_0=\operatorname{KaimingUniformInit}(\Phi)$)}}
    \KwInput{$m_0 \in \mathbb{R}^d$ \tcp{initial pre-sigmoid mask values (default: $m_0=-10\cdot\mathbf{1}_d$)}}
    \KwInput{$\lambda\in\mathbb{R}_+$ \tcp{L1-norm penalty on mask}}
    \KwOutput{$m\in\mathbb{R}^d$ \tcp{adversarial mask}}
    \BlankLine
    \SetKwProg{proc}{Procedure}{}{}
    \proc{\ApplyMask{$x\in\mathbb{R}^d$, $m\in\mathbb{R}^d$}}{
        Draw random noise $\varepsilon$ from $\mathcal{N}_d(0, I)$ \;
        Soft-threshold mask $\tilde{m} \gets \operatorname{Sigmoid}(m)$ \;
        Compute masked example $\tilde{x} \gets (\mathbf{1}_d - \tilde{m}) \odot x + \tilde{m} \odot \varepsilon$ \;
        \Return $\tilde{x}$
    }
    \BlankLine
    $t \gets 0$ \;
    \While{\upshape{not converged}}{
        Draw datapoint $(x_t, z_t)$ from $\setfnt{D}$ \;
        Compute classifier gradient $g_t^{\theta} \gets \nabla_\theta\left[  \ell\left(\Phi(\ApplyMask(x_t, m_t); \theta), z_t\right) \right]_{\theta=\theta_t}$ \;
        Update classifier parameters $\theta_{t+1} \gets \operatorname{UpdateParams}^\theta(g_t^\theta)$ \;
        Compute mask gradient $g_t^m \gets \nabla_m \left[ \lambda\lVert m \rVert_1 -\ell\left(\Phi(\ApplyMask(x_t, m); \theta_{t+1}), z_t\right) \right]_{m=m_t}$\;
        Update mask parameters $m_{t+1} \gets \operatorname{UpdateParams}^m(g_t^m)$ \;
        $t \gets t+1$ \;
    }
    \Return $\operatorname{Sigmoid}(m_t)$
    
\caption{Our adversarial mask (AdvMask) computation}
\label{alg:advmask}
\end{algorithm}

As mentioned previously, first-order statistical methods are unable to detect dependencies between sensitive variables and multiple power trace timesteps when the sensitive variable is pairwise-independent of each timestep, as is the case when Boolean masking is used to break a variable $z$ into two statistically-independent shares $r$ and $z \xor r$. The reasons for this are that 1) such methods look for dependence between $z$ and each individual timestep but not between $z$ and sets of multiple timesteps, and 2) many such methods cannot detect nonlinear dependencies between $z$ and timesteps of the power trace, such as the XOR function $z = r \xor (z \xor r)$ which must be computed to extract $z$ from the shares. In contrast, DNN-based methods are capable of detecting these dependencies due to the ability of DNNs to learn arbitrary continuous functions on $\mathbb{R}^d$. Their ability to overcome Boolean masking without assuming knowledge of masks at profiling time has been demonstrated empirically \cite{EPRINT:MagPorPro16, JCEng:BPSCD20}.

While neural network attribution methods don't suffer from the above limitation, they are limited insofar as there is no guarantee that the DNN they are applied to has absorbed every useful relationship between power traces and the target sensitive variables. Adversarial masking overcomes this issue by jointly training a classifier with an adversarial noise generator, in such a way that noise is gradually added to the timesteps which are most-used by the classifier. Over time, the classifier is thereby forced to learn new input-output relationships to retain its performance.

\subsection{Synthetic power traces}
\label{sec:power_traces}

In order to evaluate the efficacy of our method in a controlled setting and to observe the effect of various interventions on dataset properties, we have run a variety of experiments on synthetic power traces.

Our basic procedure for generating synthetic datasets is shown in \refalg{alg:synthetic_data_generation}, based on the Hamming weight power consumption model \cite{mangard2008}. We construct each power trace $x^{(i)}$ as a sum of 3 components:
\begin{equation}
    x^{(i)} = \varepsilon^{(i)}_{\text{lkg}} + \varepsilon^{(i)}_{\text{rand}} + \varepsilon_{\text{op}}
\end{equation}
where
\begin{itemize}
    \item $\varepsilon^{(i)}_{\text{lkg}}$ denotes the attack point-dependent power consumption. We model this as an affine function of the Hamming weight of the attack point at the leaking timesteps, and $0$ elsewhere. We generally assume that leakage is \textit{temporally-sparse}, i.e. most elements of this vector will be $0$.
    \item $\varepsilon^{(i)}_{\text{rand}}$ denotes the `random' component of power consumption stemming from factors which are irrelevant to carrying out a side-channel attack (e.g. electrical noise, data-dependent power consumption for attack point-irrelevant data). We model this as Gaussian noise.
    \item $\varepsilon_{\text{op}}$ denotes the operation-dependent power consumption which is the same for every encryption. We model this as Gaussian noise which is sampled once prior to dataset generation, then added to every generated power trace.
\end{itemize}
We additionally constrain the variance of the power trace be the same at every timestep (hence the expression in \refalgln{algln:data_dependent_power_consumption} of \refalg{alg:synthetic_data_generation}). The two critical assumptions are that 1) leakage is proportional to the Hamming weight of the attack point, and 2) leakage is temporally-sparse. Both are likely to approximately hold for unprotected devices which leak primarily due to power consumption when the attack point value is loaded onto a bus or a register \cite{mangard2008}. We will shortly modify \refalg{alg:synthetic_data_generation} to simulate masking and hiding countermeasures which are common in protected cryptographic implementations.

Unless stated otherwise, subsequent experiments will use the default parameter values listed in \refalg{alg:synthetic_data_generation}.

\begin{algorithm}
\KwInput{$N\in\mathbb{Z}_+$ \tcp{number of datapoints to generate (default: $N=5\times 10^4$)}}
\KwInput{$d\in\mathbb{Z}_+$ \tcp{timesteps per power trace (default: $d=128$)}}
\KwInput{$\sigma_{\text{op}}^2, \sigma_{\text{rand}}^2 \in\mathbb{R}_+$ \tcp{variance of resp. operation-dependent noise and random noise (default: $\sigma_{\text{op}}^2 = 1, \sigma_{\text{rand}}^2 = 0.5$)}}
\KwInput{$\tau_1, \dots, \tau_\ell\in\zint{0}{d-1}$ \tcp{leaking timesteps (default: 1 leaking timestep with $\tau_1 = \left\lfloor \frac{1}{2}d \right\rfloor$)}}
\KwInput{$\alpha_1, \dots, \alpha_\ell \in [0, 1]$ \tcp{proportion of variance due to attack point Hamming weight at leaking timesteps (default: each $\alpha_i = \frac{1}{2}$)}}

\KwOutput{$\setfnt{D} = \seq{(x^{(i)}, z^{(i)})}_{i=1}^{N} \in \left(\mathbb{R}^d\times\zint{0}{255}\right)^N$ \tcp{Synthetic dataset of (power trace, attack point) pairs}}
\BlankLine
Sample $\varepsilon_{\text{op}}$ from $\mathcal{N}_d(0, \sigma_{\text{op}}^2 I)$ \tcp{operation-dependent noise added to every power trace}
\BlankLine
\For{$i=1, \hdots, N$}
{
    Sample $z^{(i)}$ from $\mathcal{U}\zint{0}{255}$ \tcp{attack point value} \label{algln:sample_attack_point}
    Sample $\varepsilon_{\text{rand}}^{(i)}$ from $\mathcal{N}_d(0, \sigma_{\text{rand}}^2 I)$ \tcp{per-trace random noise}
    Compute $\varepsilon_{\text{lkg}}^{(i, j)} \gets \sqrt{\frac{\alpha_j}{2}}\left(\operatorname{HammingWeight}(z^{(i)})-4\right) + \sqrt{1-\alpha_j}(\varepsilon_{\text{rand}}^{(i)})_j$ for $j=1, \dots, \ell$ \tcp{data-dependent power consumption at leaking points} \label{algln:data_dependent_power_consumption}
    Compute $x^{(i)} \gets \varepsilon_{\text{op}} + \left( \begin{array}{ll} \varepsilon_{\text{lkg}}^{(i, k)} & \text{if}\;\;j=\tau_k\;\;\text{for some}\;\;k\in\zint{1}{\ell} \\ (\varepsilon_{\text{rand}}^{(i)})_j & \text{else} \end{array} \right)_{j=1}^{d}$ \tcp{power trace corresponding to $z^{(i)}$} \label{algln:trace_done}
}
\Return $\setfnt{D} = \seq{(x^{(i)}, z^{(i)})}_{i=1}^{N}$
\caption{Basic synthetic dataset generation}
\label{alg:synthetic_data_generation}
\end{algorithm}

\subsection{Evaluation metrics}
\label{sec:eval_metrics}

Given a mask $m \defeq (m_i)_{i=1}^{d}\in [0, 1]^d$ and a dataset $\setfnt{D}$ generated by \refalg{alg:synthetic_data_generation}, we can define the index sets
\begin{equation}
    \setfnt{I}_{\text{lkg}} \defeq \{\tau_1, \dots, \tau_\ell\} \;\;\text{and}\;\; \setfnt{I}_{\lnot\text{lkg}} \defeq \{1, \dots, d\} \setminus \setfnt{I}_{\text{lkg}}.
\end{equation}
We will summarize the quality of $m$ with respect to $\setfnt{D}$ using the mean z-score
\begin{equation}
    \zscore(m; \setfnt{I}_{\text{lkg}}, \setfnt{I}_{\lnot\text{lkg}}) \defeq \avg_{i\in\setfnt{I}_{\text{lkg}}} \left[\frac{m_i - \avg_{j\in\setfnt{I}_{\lnot\text{lkg}}} m_j}{\std_{j\in\setfnt{I}_{\lnot\text{lkg}}}m_j}\right]
\end{equation}
and the area under the precision-recall curve, where we define precision
\begin{equation}
    \operatorname{Precision}(m; z, \setfnt{I}_{\text{lkg}}, \setfnt{I}_{\lnot\text{lkg}}) \defeq \frac{|\{\tau=1, \dots, d : m_\tau \geq z, \tau \in \setfnt{I}_{\text{lkg}}\}|}{|\{\tau=1, \dots, d : m_\tau \geq z\}|}
\end{equation}
and recall
\begin{equation}
    \operatorname{Recall}(m; z, \setfnt{I}_{\text{lkg}}, \setfnt{I}_{\lnot\text{lkg}}) \defeq \frac{|\{\tau=1, \dots, d : m_\tau \geq z, \tau \in \setfnt{I}_{\text{lkg}}\}|}{|\{\tau=1, \dots, d : m_\tau \geq z, \tau \in \setfnt{I}_{\text{lkg}}\} \cup \{\tau=1, \dots, d : m_\tau < z, \tau \in \setfnt{I}_{\text{lkg}}\}|},
\end{equation}
then numerically approximate the area under the curve with recall on the horizontal axis and precision on the vertical axis as we sweep the threshold $z$ from $0$ to $1$.

We will see that early stopping is necessary to select a high-quality mask when training for many epochs with both the GradVis and AdvMask algorithms. Given a set of masks $\setfnt{M} \subset [0,1]^d$ generated by e.g. recording the mask at the end of each epoch of training, we will select a `best' mask $m^*\in \setfnt{M}$ using the policy
\begin{equation}
    m^* \in \arg\max_{m \in \setfnt{M}'} \zscore(m; \setfnt{I}_{\text{lkg}}, \setfnt{I}_{\lnot\text{lkg}}) \;\;\text{where}\;\; \setfnt{M}' \defeq \arg\max_{m\in\setfnt{M}} \operatorname{PR-AUC}(m; \setfnt{I}_{\text{lkg}}, \setfnt{I}_{\lnot\setfnt{lkg}}).
\end{equation}
While this policy cannot be used in practice because it relies on `oracle' knowledge of which timesteps leak the key, it provides a useful basis for comparing mask-generation algorithms on synthetic datasets for which the leaking timesteps are known.

\section{Results}

\subsection{Implementation details and hyperparameter settings}
\label{sec:implementation}

We have implemented the following experiments in PyTorch \cite{paszke2019}, with some evaluation metrics computed using scikit-learn \cite{pedregosa2011}. Our code can be found \href{https://github.com/jimgammell/gan_side_channel_leakage_detector/tree/main}{here}.  %\href{https://github.com/jimgammell/gan_side_channel_leakage_detector/tree/main}{here}.

\begin{figure}
    \centering
    \includegraphics[width=\textwidth]{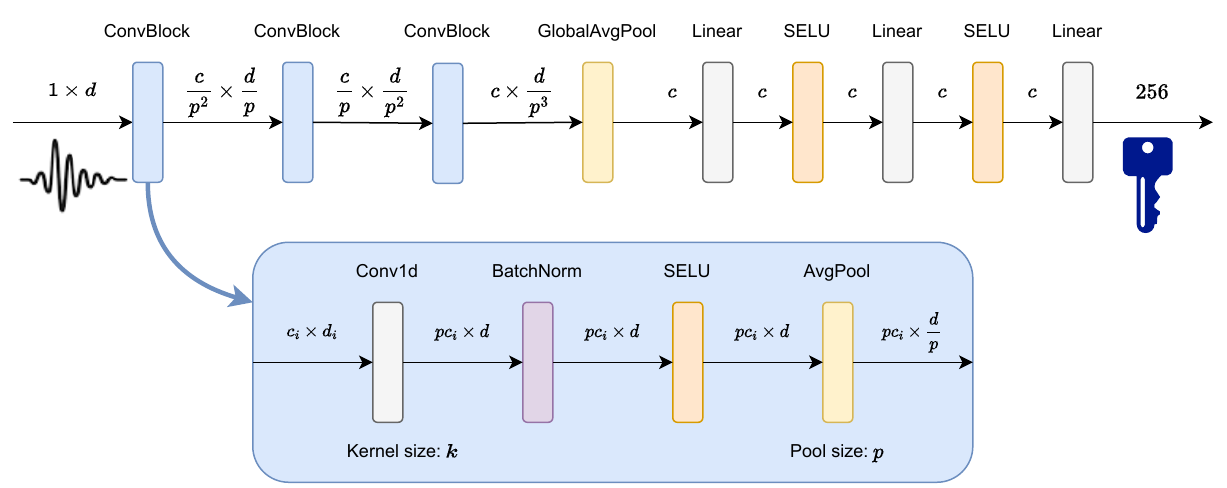}
    \caption{Convolutional neural network classifier architecture used in experiments.}
    \label{fig:specific_cnn}
\end{figure}

Unless otherwise specified, in the following experiments we use the CNN architecture described in \reffig{fig:specific_cnn} for all classifiers. The architecture style based on convolution--activation--pooling blocks followed by linear--activation blocks was popular in early applications of DNNs to computer vision \cite{lecun1998, krizhevsky2012, simonyan2015} and has proven successful for side-channel analysis of short power traces \cite{JCEng:BPSCD20, EPRINT:ZBHV19, TCHES:WAGP20}. We use a global average pooling (GlobalAvgPool) layer \cite{lin2013} after the convolutional blocks rather than the typical flattening layer, so that we can apply the architecture to arbitrary-length inputs without a quadratic increase in parameter count. As was done in \cite{EPRINT:ZBHV19}, we use batch normalization (BatchNorm) \cite{ioffe2015} and the scaled exponential linear unit (SELU) activation \cite{klambauer2017}. We train all models with the Adam \cite{kingma2015} variant of minibatch stochastic gradient descent. For each dataset we compute the mean and standard deviation of all timesteps of all traces and use these to standardize the traces:
\begin{equation}\begin{gathered}
    x^{(i)} \gets \left( \frac{x^{(i)}_j - \mu}{\sigma} \right)_{j=1}^{d} \\
\text{where}\;\;
    \mu \defeq \avg_{\substack{i\in\zint{1}{N} \\ j\in\zint{1}{d}}} x^{(i)}_j \;\;\text{and}\;\;\sigma\defeq\std_{\substack{i\in\zint{1}{N{}} \\ j\in\zint{1}{d}}} x^{(i)}_j.
\end{gathered}\end{equation}
We do not standardize the individual timesteps of each trace (i.e. compute per-timestep means and standard deviations $\mu, \sigma \in \mathbb{R}^d$) because the global average pooling layer removes information about the absolute position of timesteps, forcing the network to rely on relative position information which is encoded in the mean values of neighboring timesteps (i.e. the operation-dependent power consumption).

Unless otherwise specified, we use the hyperparameter values listed in \reftbl{tab:hparams}. These have been tuned in an \textit{ad hoc} manner and seem to work in a variety of settings, but are not necessarily the optimal values for any particular setting. For each trial we use the same classifier architecture and classifier optimizer settings for both the GradVis and AdvMask methods. We compute the SNR using the full dataset, in contrast to the DNN-based methods where we partition the dataset into a training and validation subset and do not train on the latter.

Our goal with these experiments is to compare the trends and failure modes of each of the 3 considered methods in various settings. They do not conclusively demonstrate that AdvMask is `better' than GradVis, and to do so it would be necessary to independently tune the hyperparameters for both methods under a fixed computational budget for each trial, and to explore a wider variety of hyperparameters and neural network architectures to ensure we have not settled in a local minimum of the search space with respect to one method, which is sub-optimal for the other.

\renewcommand{\arraystretch}{1.25}
\begin{table}[]
    \centering
    \begin{tabular}{c|c}
        \textbf{Hyperparameter} & \textbf{Default value} \\\hline\hline
        Classifier architecture & \reffig{fig:specific_cnn} with $c=16$, $p=2$, $k=11$ \\\hline
        Classifier optimizer & Adam($\alpha=10^{-3}$, $\beta_1=0.9$, $\beta_2=0.999$, $\epsilon=10^{-8}$) \\\hline
        Mask optimizer & Adam($\alpha=10^{-2}$, $\beta_1=0.5$, $\beta_2=0.0$, $\epsilon=10^{-8}$) \\\hline
        Mask L1 norm penalty $\lambda$ & 1.0 \\\hline
        Minibatch size & 256 \\\hline
        Validation split proportion & 0.2\\\hline
        Training duration & $10^4$ minibatches
    \end{tabular}
    \caption{Hyperparameter settings used in experiments, except where specified otherwise.}
    \label{tab:hparams}
\end{table}

\subsection{Experiments on synthetic data}

Here we evaluate the performance of our AdvMask technique on synthetic power trace datasets generated in the manner described in section \refsec{sec:power_traces}. We compare it to the SNR (\refsec{sec:snr}) and GradVis (\refsec{sec:gradvis}) techniques as representative examples of techniques based on resp. first-order statistical calculations and neural network attribution. We repeat L1 norm sweeps for 3 random seeds and dataset setting sweeps for 5 random seeds. In the subsequent plots, shaded areas cover the full range of values measured across these repeated trials, with dots denoting the median values and dotted lines interpolating the dots.

\subsubsection{Impact of proportion of variance due to leakage}

We first examine the impact of the proportion of variance due to leakage at the leaking points. We construct datasets according to \refalg{alg:synthetic_data_generation}, with all inputs set to their default values except that we sweep the proportion of variance due to the Hamming weight of the target variable, $\alpha_1$, from 0 to 1.

\begin{figure}
    \centering
    \includegraphics[width=\textwidth]{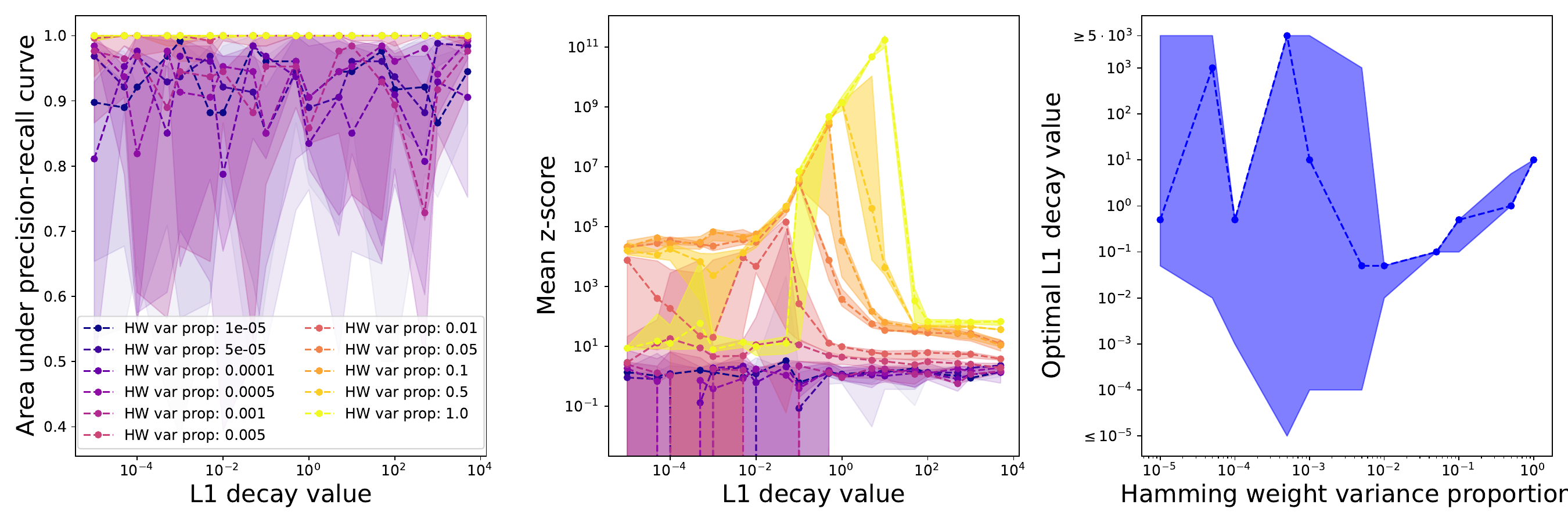}
    \caption{Sweep of the AdvMask L1 norm penalty, while varying the proportion of variance at leaking points due to the Hamming weight of the sensitive variable. Observe that for sufficiently-large variance, the optimal L1 decay value increases with variance. When the variance is very low, AdvMask fails to identify the leaking point and the optimal L1 norm penalty is volatile.}
    \label{fig:hw_var_prop_l1}
\end{figure}

In \reffig{fig:hw_var_prop_l1} we examine the impact of $\alpha_1$ on the optimal mask L1 norm penalty $\lambda$. We see that for $\alpha_1 < 10^{-2}$ when there is very little leakage, our technique fails to detect the leaking point and the optimal value of $\lambda$ is volatile. For $\alpha_1 \geq 10^{-2}$, our method starts to work and we see that the optimal $\lambda$ increases with $\alpha_1$.

\begin{figure}
    \centering
    \includegraphics[width=\textwidth]{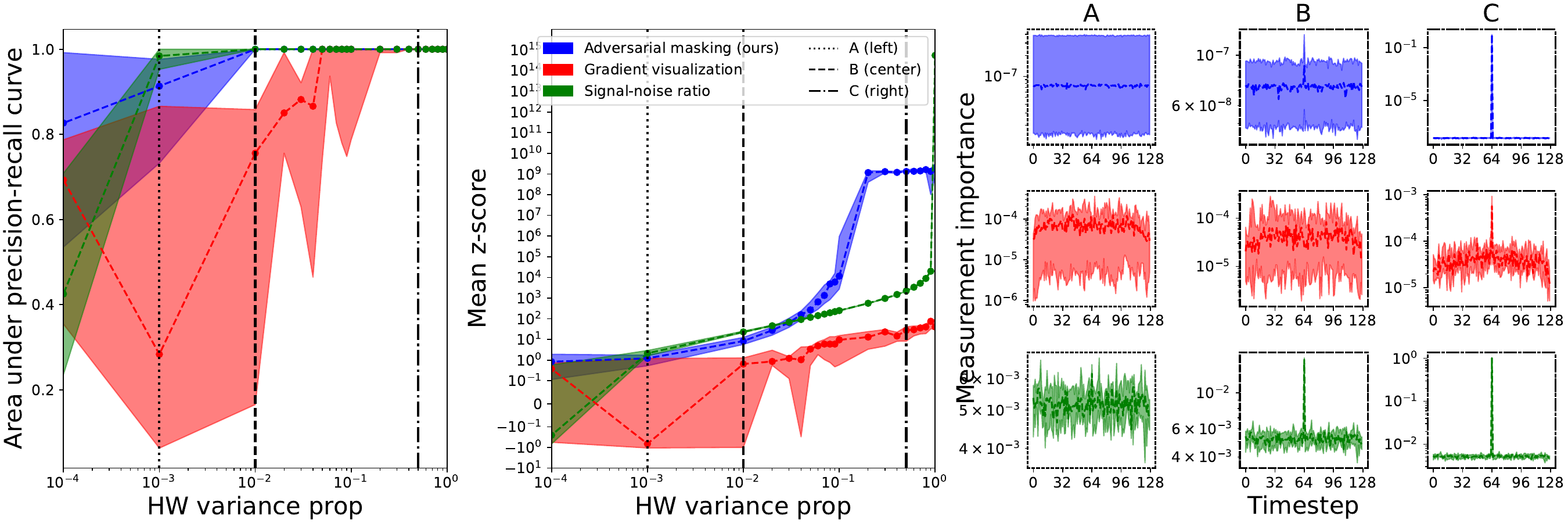}
    \caption{Sweep of the proportion of variance at leaking points due to the Hamming weight of the sensitive variable. Observe that AdvMask (ours) and SNR have similar minimal variance requirements, while GradVis requires significantly-more variance to identify the leaking timestep.}
    \label{fig:vs_noise}
\end{figure}

In \reffig{fig:vs_noise} we compare AdvMask to SNR and GradVis as the amount of dataset noise varies. We see that all 3 methods fail with $\alpha_1$ is extremely low and succeed when it grows sufficiently large. AdvMask and SNR are comparably-robust to $\alpha_1$, with the former outperforming the latter at moderate $\alpha_1$, likely due to the inductive bias imposed by the L1 norm penalty in its training routine. GradVis performs the worst, failing to detect the leaking point until $\alpha_1 = 2\times 10^{-1}$; we suspect it underperforms AdvMask because the noise added in the AdvMask training procedure provides a useful regularization effect which prevents overfitting.

\subsubsection{Impact of number of distinct points which are leaking}

Next, we vary the number of points per power trace which leak the key, setting $\tau_i = \left\lfloor \frac{i}{n+1}d \right\rfloor$ for $i=1, \dots, n$ while varying $n$. In the first experiment we also set the trace length to $d=512$, while in the second experiment we leave all other inputs to \refalg{alg:synthetic_data_generation} at their default values.

\begin{figure}
    \centering
    \includegraphics[width=\textwidth]{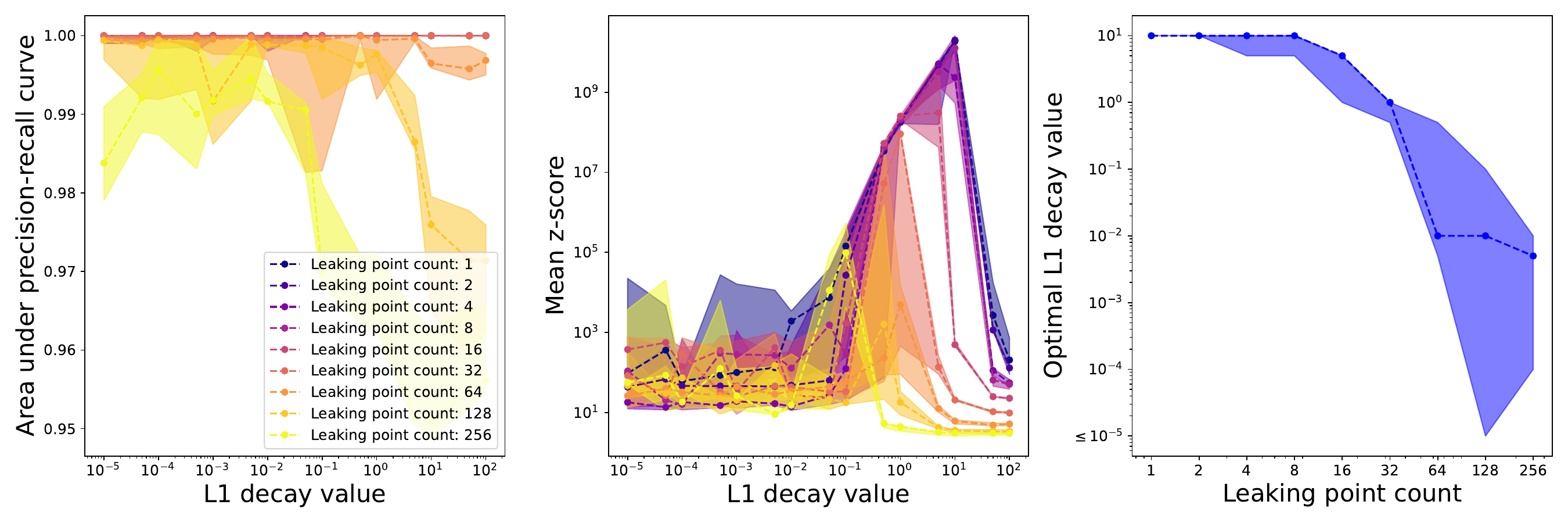}
    \caption{Sweep of the AdvMask L1 norm penalty, while varying the number of leaking points. Observe that the optimal norm penalty decreases as the number of leaking points increases.}
    \label{fig:l1_sweep__num_leaking_points}
\end{figure}

In \reffig{fig:l1_sweep__num_leaking_points} we examine the impact of $n$ on the optimal L1 norm penalty $\lambda$. We see that as we increase $n$, the optimal value of $\lambda$ decreases, which makes sense because $\lambda$ determines the tradeoff the mask should make between increasing classifier loss and minimizing the amount of added noise, and as a greater proportion of points leaks, the amount of added noise should increase.

\begin{figure}
    \centering
    \includegraphics[width=\textwidth]{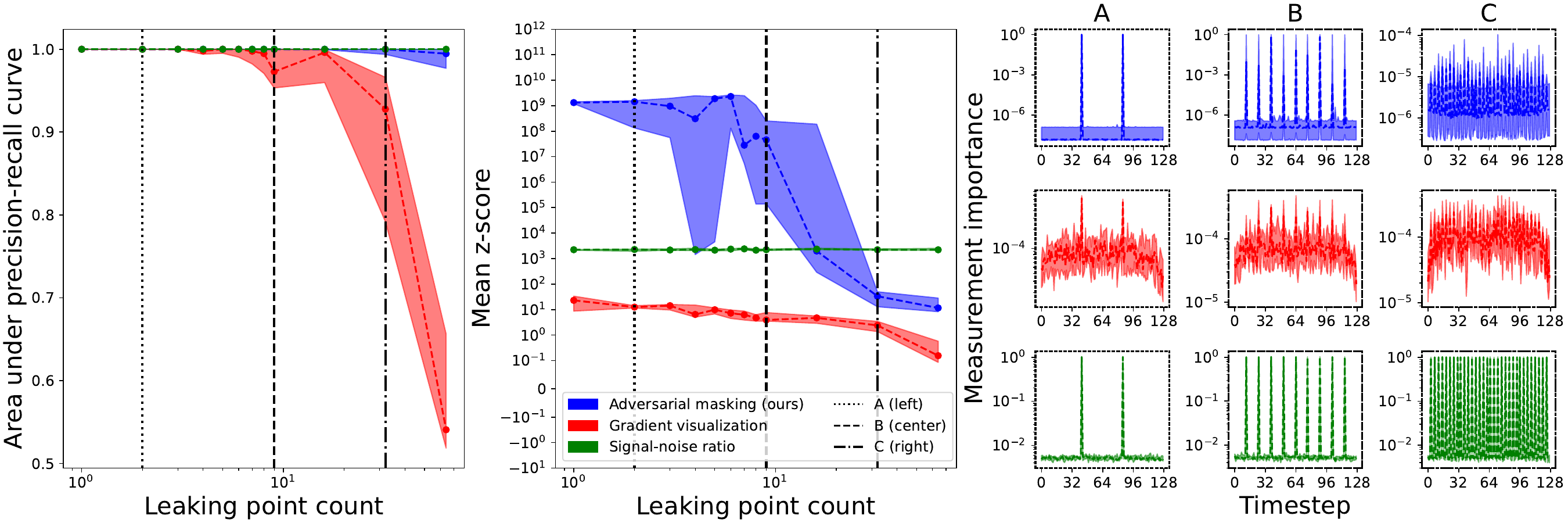}
    \caption{Sweep of the number of timesteps at which the key leaks. SNR is unaffected by the number of leaking points, whereas AdvMask (ours) sees a modest performance degradation and GradVis sees a very-substantial performance degradation.}
    \label{fig:vs_num_leaking_points}
\end{figure}

In \reffig{fig:vs_num_leaking_points} we compare AdvMask to SNR and GradVis as the number of leaking points varies. We see that the number of leaking points has no impact on the SNR, as it computes the statistics of each leaking point in isolation. Both AdvMask and GradVis seek reduced performance as $n$ increases, but the performance degradation for GradVis is much stronger, as it begins to miss a substantial proportion of leaking points when $n\geq 64$. We believe AdvMask performs better in this setting because while in the GradVis setting the classifier can accurately predict the sensitive variable using only a small number of leaking points, the simultaneous adversarial training of the classifier and noise generator in the AdvMask setting forces the classifier to use as many points as possible as those it currently relies are progressively attenuated by the noise generator.

\subsubsection{Impact of power trace length}

Here we vary the number of timesteps per power trace $d$, while leaving other inputs to \refalg{alg:synthetic_data_generation} at their default values.

\begin{figure}
    \centering
    \includegraphics[width=\textwidth]{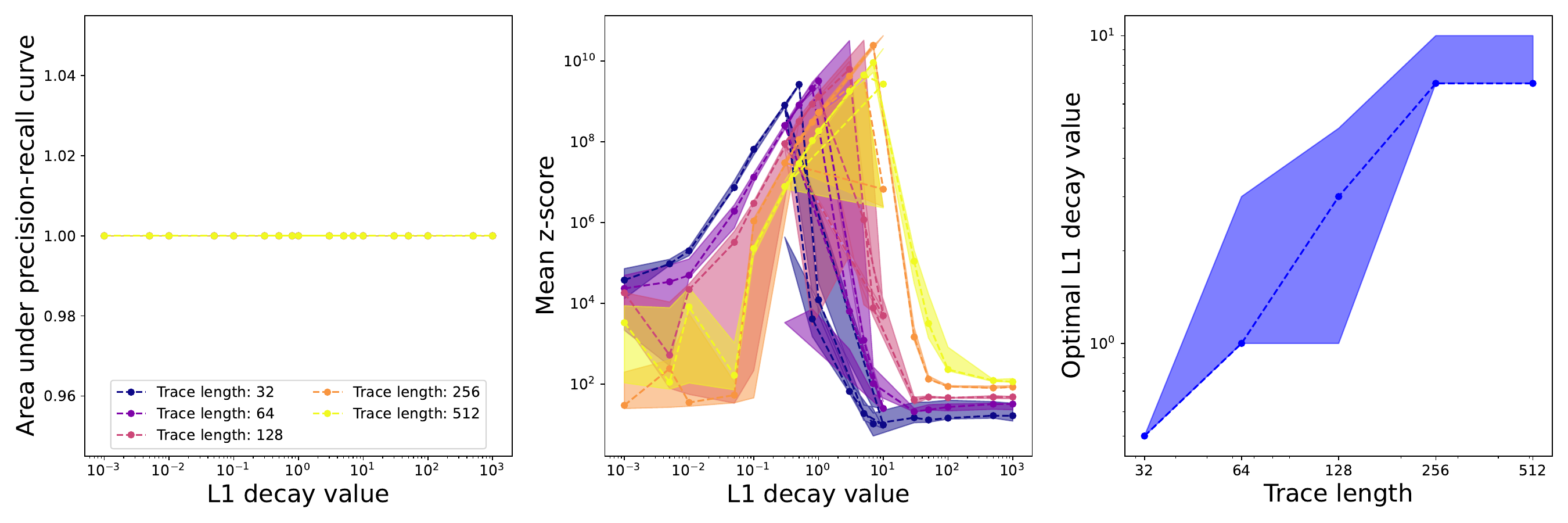}
    \caption{Varying-length power traces, sweeping the AdvMask norm penalty. Observe that the optimal penalty increases with trace length.}
    \label{fig:trace_length_l1_sweep}
\end{figure}

In \reffig{fig:trace_length_l1_sweep} we examine the impact of $d$ on the optimal L1 norm penalty $\lambda$. We see that trace length does not substantially change results, with every $\lambda$ successfully detecting the leaking point, though there is a small increase in the optimal $\lambda$ as traces become longer.

\begin{figure}
    \centering
    \includegraphics[width=\textwidth]{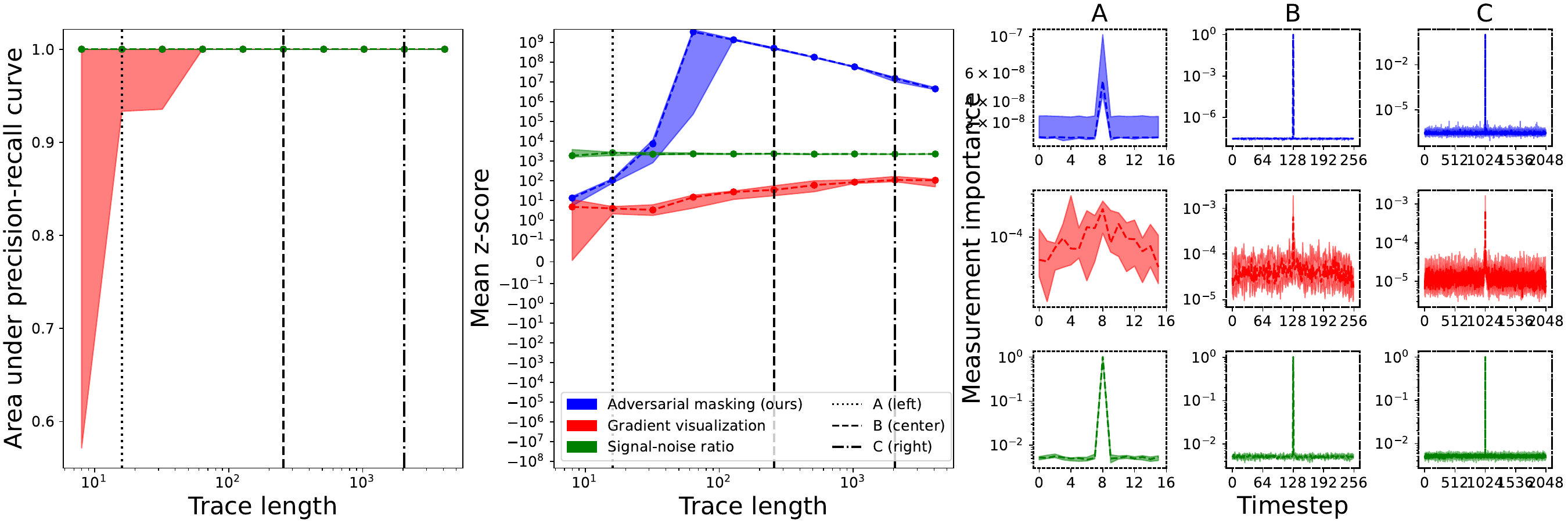}
    \caption{Sweep of the number of timesteps per power trace. SNR is unaffected by power trace length, whereas AdvMask (ours) and GradVis see some performance degradation for very-short power traces.}
    \label{fig:trace_length_sweep}
\end{figure}

In \reffig{fig:trace_length_sweep} we compare AdvMask to SNR and GradVis as the length of power traces increases. As expected, the length of the power trace has no impact on the performance of SNR, which considers timesteps in isolation. Surprisingly, both AdvMask and GradVis see some performance degradation when the lengths of power traces are very small. One possible explanation is that the CNN classifier architecture and optimizer hyperparameters used by both GradVis and AdvMask are inappropriate for small power traces. For example, convolutional layers offer a regularizing effect when the power trace length is substantially larger than the convolutional kernel, but become increasingly similar to linear layers as power trace length decreases.

\subsubsection{Impact of trace desynchronization}

Here we add random delays to the power traces to simulate trace desynchronization which may stem from e.g. random delays or clock jitter inserted into the AES implementation. To implement random delays of duration at most $m$ timesteps, we execute \refalg{alg:synthetic_data_generation} to generate traces of length $d+m$, then for each trace $x^{(i)}$ after \refalgln{algln:trace_done} crop out a random $d$-length sub-trace:
\begin{equation}
    x^{(i)} \gets \left( x^{(i)}_j \right)_{j=m^{(i)}+1}^{d+m^{(i)}} \;\;\text{for}\;\; m^{(i)} \sim \mathcal{U}\zint{0}{m}.
\end{equation}
In the subsequent experiments we vary $m$ while leaving the other inputs to \refalg{alg:synthetic_data_generation} at their default values.

\begin{figure}
    \centering
    \includegraphics[width=\textwidth]{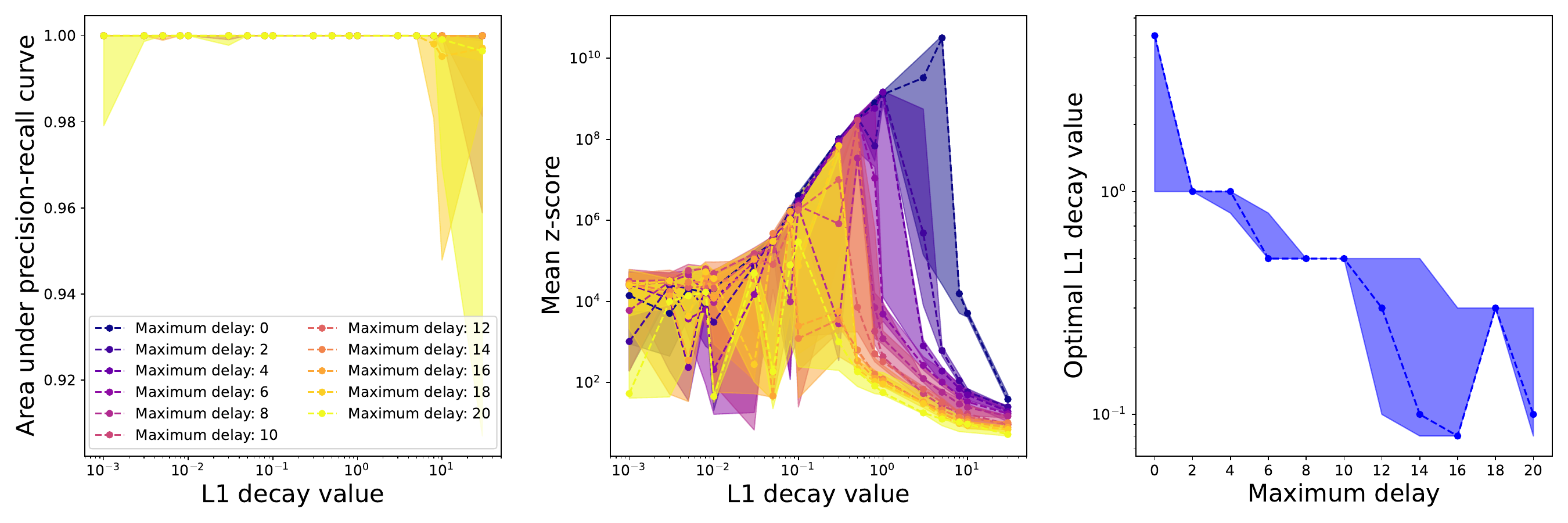}
    \caption{Datasets with randomly desynchronized traces, sweeping the AdvMask L1 norm penalty. Observe that the optimal value decreases as the amount of desynchronization increases.}
    \label{fig:trace_desynchronization_l1}
\end{figure}

In \reffig{fig:trace_desynchronization_l1} we examine the impact of trace desynchronization on the optimal value of the mask L1 norm penalty $\lambda$. We see that the optimal value of $\lambda$ decreases significantly as $m$ increases. Trace desynchronization increases the number of leaking timesteps, so this effect makes sense given that increasing the number of timesteps which leak also decreases the optimal value of $\lambda$.

\begin{figure}
    \centering
    \includegraphics[width=\textwidth]{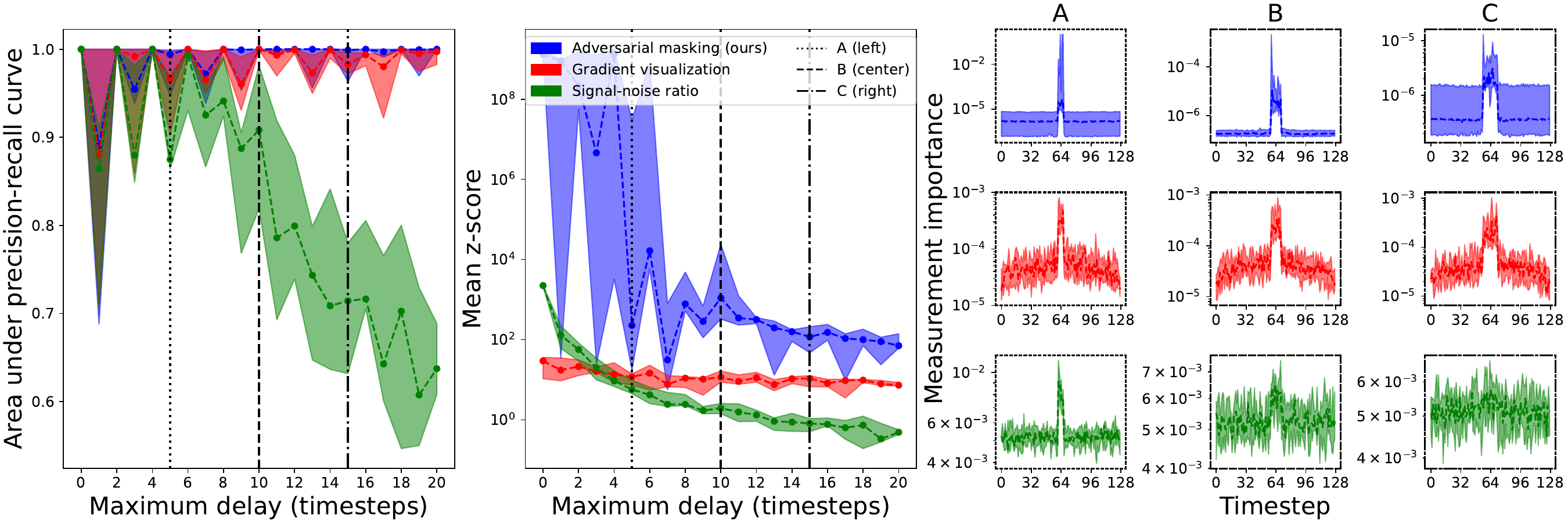}
    \caption{Dataset with randomly desynchronized traces, sweeping the maximum amount of desynchronization. Observe that the performance of SNR quickly drops off, whereas both AdvMask (ours) and GradVis are robust to desynchronization due to their use of a convolutional neural network for leakage localization. }
    \label{fig:vs_trace_desynchronization}
\end{figure}

In \reffig{fig:vs_trace_desynchronization} we compare AdvMask to SNR and GradVis as the amount of desynchronization increases. Observe that while AdvMask and GradVis see a small performance degradation as the amount of desynchronization increases, they remain performant, detecting most of the leaking points. In contrast, SNR sees substantial performance degradation. This highlights one of the major advantages of deep learning leakage localization methods: it is possible to modify the DNN architecture or objective function to impose inductive biases towards functions which are expected \textit{a priori} to generalize well. In this case, the convolution and pooling layers in the classifier architectures of AdvMask and GradVis bias them towards functions which are invariant to temporal shifts of the input, which makes them robust to trace desynchronization.

\subsubsection{Impact of Boolean masking}

Here we simulate Boolean masking, which is a common countermeasure to increase the difficulty of discerning sensitive variables by breaking them up into statistically-independent shares which are operated on at distant points in time. In \refalgln{algln:sample_attack_point} of \refalg{alg:synthetic_data_generation} we sample both an attack point value $z^{(i)}$ and a Boolean mask value $r^{(i)}$ from $\mathcal{U}\zint{0}{255}$, then compute a `masked' attack point $z^{(i)} \xor r^{(i)}$ where $\cdot\xor\cdot$ denotes the bitwise XOR operation. We then implement leakage for both $r^{(i)}$ and $z^{(i)} \xor r^{(i)}$ at separate timesteps in the same manner as above.

Observe that $z^{(i)} \xor r^{(i)}$ is independent of $z^{(i)}$ unless information about $r^{(i)}$ is known. This makes $z^{(i)}$ more-challenging to attack because to learn its value one must learn the values of both $r^{(i)}$ and $z^{(i)} \xor r^{(i)}$ which are leaked at distant timesteps, then perform the nonlinear computation $r^{(i)} \xor (z^{(i)} \xor r^{(i)}) = z^{(i)}.$ Because two values must be learned this is often called \textit{second-order} leakage, while leakage in the manner described above is called \textit{first-order} leakage.

\begin{figure}
    \centering
    \includegraphics[width=\textwidth]{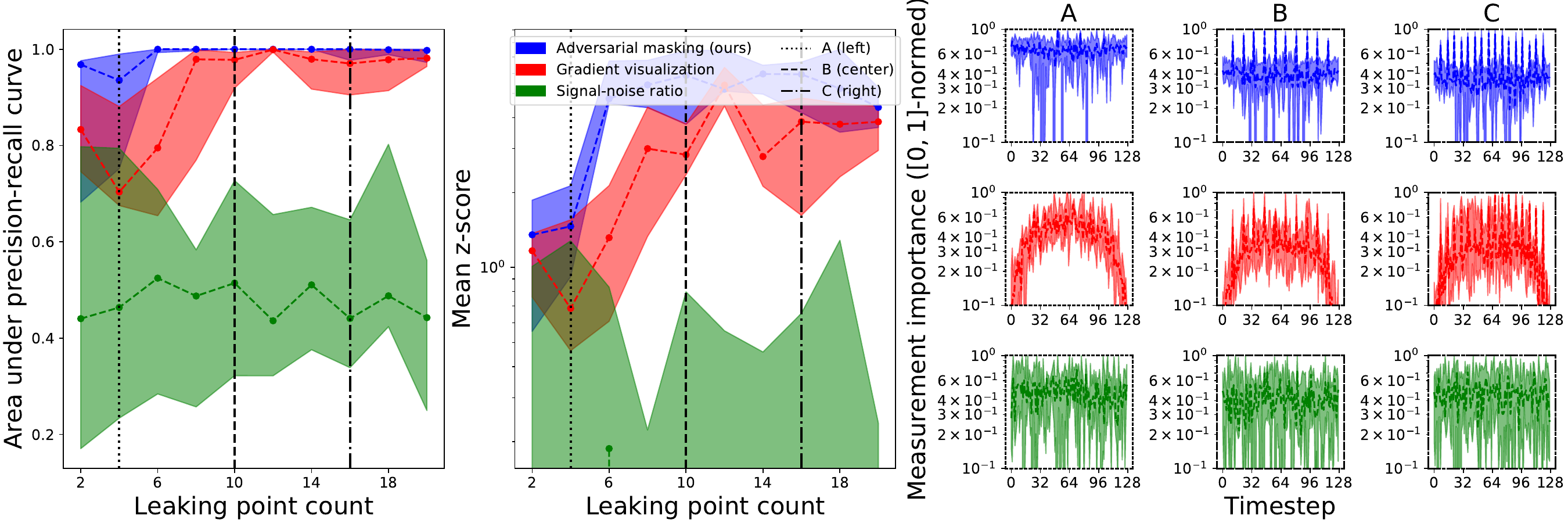}
    \caption{Boolean-masked dataset, sweeping the number of leaking points. Observe that SNR fails in all cases, and given a sufficient number of leaking points GradVis will generally identify some but not all leaking points. In contrast, AdvMask (ours) consistently identifies most or all leaking points when at least 6 are present.}
    \label{fig:vs_count_boolean_masked}
\end{figure}

\begin{figure}
    \centering
    \includegraphics[width=\textwidth]{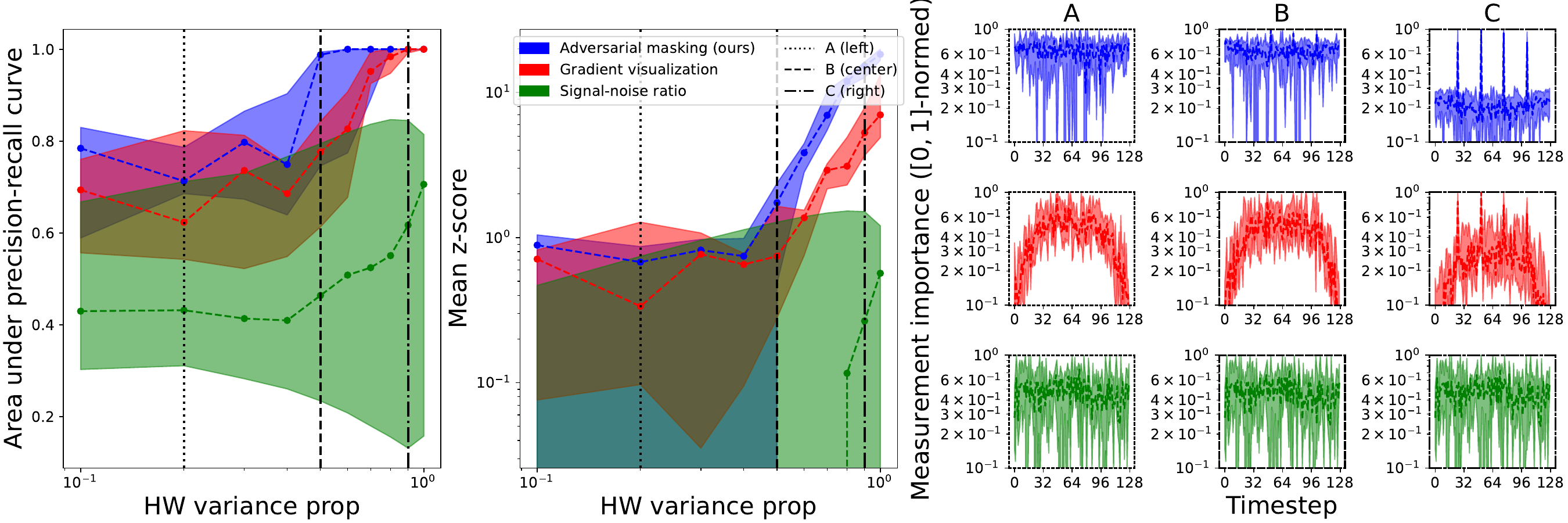}
    \caption{Boolean-masked dataset, sweeping the proportion of variance at leaking points due to the Hamming weight of the sensitive variable. Observe that SNR fails in all cases to identify second-order leaking points, whereas both AdvMask (ours) and GradVis succees when the proportion is sufficiently-large.}
    \label{fig:vs_noise_boolean_masked}
\end{figure}

In \reffig{fig:vs_count_boolean_masked} we compare AdvMask to SNR and GradVis while varying the number of timesteps at which $r$ and $r \xor z$ leak, with $\alpha_i=0.5$ for all $i$. In \reffig{fig:vs_noise_boolean_masked} we vary the proportion of variance due to $r$ and $r \xor z$ while leaving the number of leaking timesteps fixed at $4$ ($2$ for each share). Second-order leakage is harder to classify than first-order leakage, and we make the following \textit{ad hoc} changes to hyperparameter settings to increase the consistency of AdvMask and GradVis: we increase the dataset size to $N=10^5$, we increase the classifier kernel size to $k=51$, and we increase the mask L1 norm penalty to $\lambda=10$. Otherwise, hyperparameters are left at their default settings.

We see that both increasing the $\alpha_i$'s and increasing the number of leaking points improves the ability of AdvMask and GradVis to identify leaking points, and when either of the settings is sufficiently-large, both methods work despite the presence of Boolean masking. In contrast, the SNR technique can never identify the leaking points regardless of the dataset configuration. This illustrates the other major advantage of deep learning-based leakage localization strategies over those based on first-order statistics: DNNs work in the presence of Boolean masking because they can approximate arbitrary continuous functions, whereas first-order techniques are oblivious to this higher-order leakage because they consider single timesteps in isolation.

\subsection{Experiments on real datasets}

\begin{figure}
    \centering
    \includegraphics[width=\textwidth]{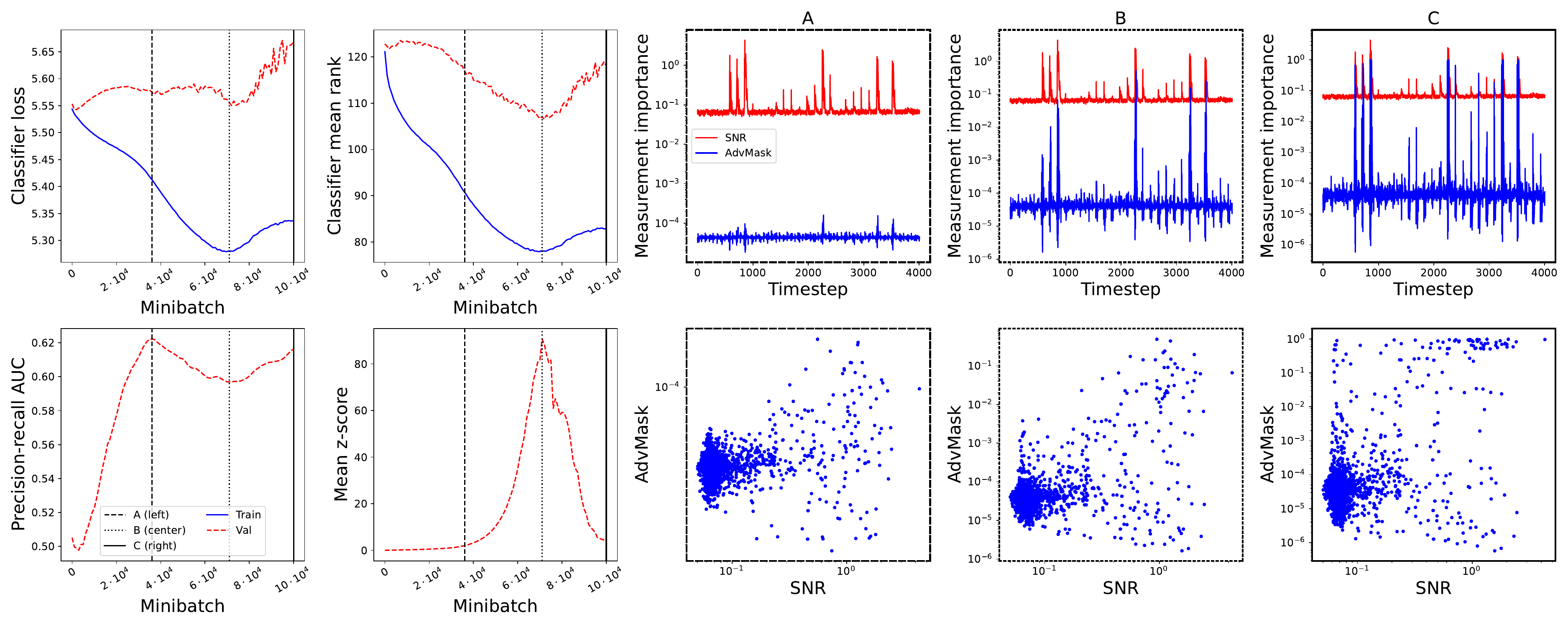}
    \caption{Results on the DPAv4 dataset. Performance on this dataset is fairly good, with AdvMask and SNR identifying similar leaking points throughout training.}
    \label{fig:dpav4_results}
\end{figure}

\begin{figure}
    \centering
    \includegraphics[width=\textwidth]{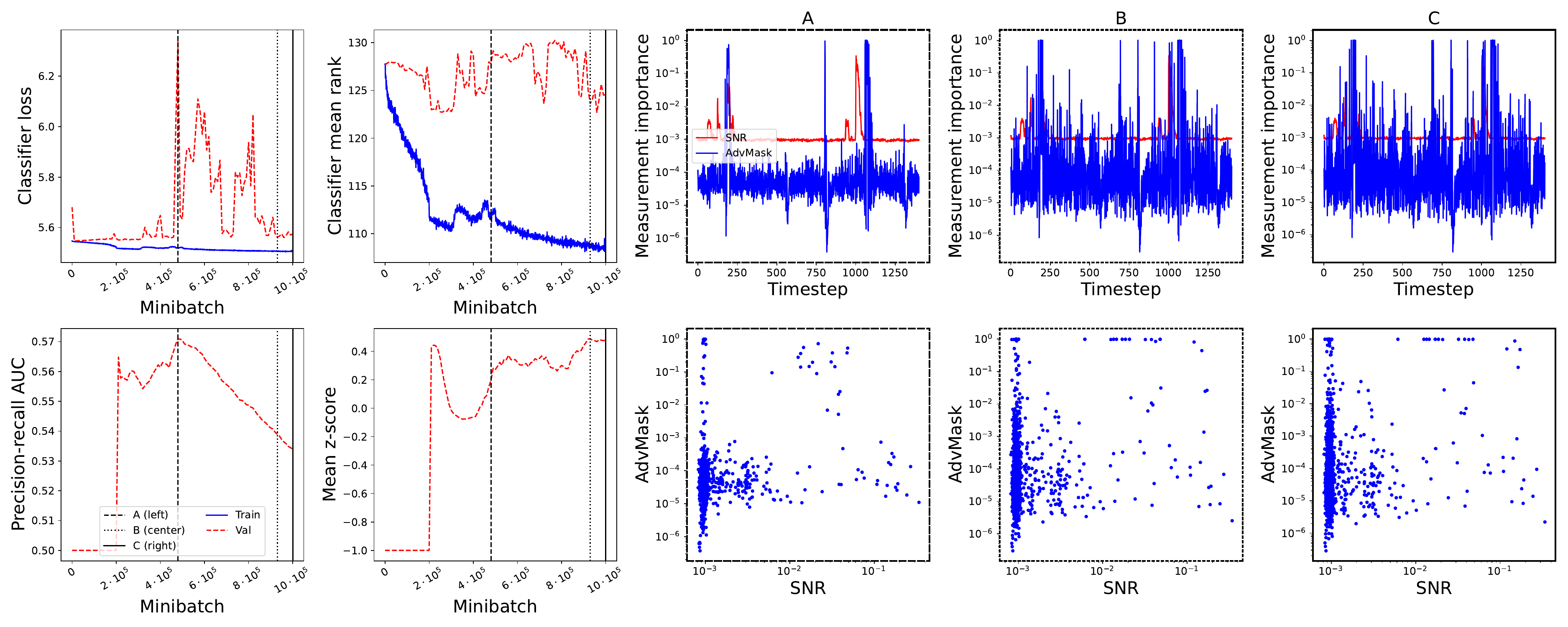}
    \caption{Results on the ASCADv1-random dataset. Performance on this dataset is highly-sensitive to classifier overfitting, and while the classifier identifies a subset of the peaks which are identified by SNR with ground truth knowledge of Boolean mask values, there are still a significant number of false-positive and false-negative leakage identifications.}
    \label{fig:ascadr_results}
\end{figure}

Here we present results from applying AdvMask to publicly-available AES power trace datasets. In \reffig{fig:dpav4_results} we apply it to the version of the DPAv4 context dataset \cite{bhasin2014} which was released alongside \cite{EPRINT:ZBHV19}, which is an unprotected AES implementation (while the implementation is masked, this version of the dataset uses the masked attack points as targets, which amounts to assuming knowledge of the mask values at both profiling-time and attack time). In \reffig{fig:ascadr_results} we apply it to the synchronized ASCADv1-random dataset \cite{JCEng:BPSCD20}, which has first-order Boolean masking.

In these experiments we cannot use the model selection strategy discussed in \refsec{sec:eval_metrics} because we do not have oracle knowledge of the leaking timesteps. Instead, we select models based on SNR measurements. Since the DPAv4 dataset is unprotected, SNR is a useful way to localize leakage and we compute the SNR with respect to the target variables. For the ASCADv1-random dataset, using the notation of \cite{JCEng:BPSCD20} we target the variable $\sbytes(p[3] \xor k[3])$, which leaks primarily through the variables $\sbytes(p[3] \xor k[3]) \xor r_{\text{out}}$, $r_{\text{out}}$, $\sbytes(p[3] \xor k[3]) \xor r[3]$, and $r[3]$, so we compute the SNR with respect to these 4 variables and take the average to get a vector $m\in\mathbb{R}^d$. For each of these datasets, we then construct `oracle' leakage index sets using the following procedure:
\begin{enumerate}
    \item Compute the SNR $\tilde{m}\in\mathbb{R}^d$ for a random dataset $\{(x_i, z_i)\}_{i=1}^{N}\in\left(\mathbb{R}^d\times\zint{0}{255}\right)^N$ where each $x_i \sim \mathcal{N}_d(0, I)$ and $z_i \sim \mathcal{U}\zint{0}{255}$ are drawn i.i.d.
    \item Fit a Gaussian distribution to $\tilde{m}$: $\mu^*, \sigma^* \in \arg\max_{\mu, \sigma\in\mathbb{R}} \prod_{i=1}^{d} \mathcal{N}(\tilde{m}_i | \mu, \sigma)$.
    \item Construct $\setfnt{I}_{\text{lkg}} \defeq \{ i=1, \dots, d : m_i \geq \mu^* + 3\sigma^* \}$ and $\setfnt{I}_{\lnot\text{lkg}} \defeq \zint{1}{d}\setminus \setfnt{I}_{\text{lkg}}$.
\end{enumerate}
Using these index sets we can then compute the mean z-score and the precision-recall AUC as described in \refsec{sec:eval_metrics}. We take model checkpoints at the maximum values of both of these metrics, as well as after the completion of training. We make various \textit{ad hoc} to the hyperparameter settings listed in \refsec{sec:implementation}; changed hyperparameters are listed in \reftbl{tab:real_dataset_hparams}.

The AdvMask results are mostly consistent with the SNR results on the DPAv4 dataset, as expected. On the ASCADv1 dataset the AdvMask method produces a large number of false-positive timesteps and a small number of false-negative ones, with respect to the SNR results. Early stopping seems to be critical in both cases, with AdvMask producing a large number of false-positive results if the classifier is trained for long enough to overfit to noise in the dataset. While we view these results as promising in the sense that AdvMask can produce qualitatively-similar results to SNR, the reliance on oracle knowledge of leaking points is a major limitation which will need to be addressed before AdvMask can be used for practical applications.

\begin{table}[]
    \centering
    \begin{tabular}{c|c|c}
         & DPAv4 & ASVADv1-random \\\hline\hline
        Classifier learning rate $\alpha$ & $10^{-5}$ & $10^{-4}$ until batch $2\times 10^5$, then $10^{-5}$ \\\hline
        Mask learning rate $\alpha$ & $10^{-4}$ & 0 until batch $2\times 10^5$, then $10^{-4}$ \\\hline
        Mask L1 norm penalty $\lambda$ & $10^{-2}$ & $10^{-4}$ \\\hline
        Batch size & 64 & 64 \\\hline
        Number of batches & $10^{5}$ & $10^6$ \\\hline
        Classifier MLP width & 64 & 64
    \end{tabular}
    \caption{Hyperparameters for the experiments on the DPAv4 and ASCADv1-random datasets. Unless specified here, hyperparameters take on the same values as stated in \refsec{sec:implementation}.}
    \label{tab:real_dataset_hparams}
\end{table}

% Omitting this because results don't look very good, and the dataset is not well-known and does not have unique properties compared w/ DPAv4.
%\begin{figure}
%    \centering
%    \includegraphics[width=\textwidth]{figures/plots/google_tinyaes_results.pdf}
%    \caption{Results on the Google TinyAES dataset.}
%    \label{fig:google_tinyaes_results}
%\end{figure}

\section{Conclusion}

We have proposed a technique for identifying the power trace timesteps which leave cryptographic implementations vulnerable to side-channel attacks, through an adversarial game played between a deep neural network (DNN)-based classifier which seeks to determine a sensitive variable from power traces, and a trainable noise generator which seeks to prevent the sensitive variable from leaking by adding as little noise as possible to the power traces. We have demonstrated on synthetic datasets that our technique has advantages over leakage localization algorithms based on first-order statistics when implementations have countermeasures such as trace desynchronization and Boolean masking. It outperforms neural network attribution algorithms when there are many sources of leakage, a strict subset of which are sufficient for accurately classifying the sensitive variable. We have also demonstrated that on the DPAv4 and ASCADv1-random datasets, our technique can produce qualitatively similar results to signal-noise ratio measurements which rely on profiling-time access to Boolean mask values.

A major barrier to practical usage of our technique is that early stopping is essential to prevent false-positive leakage identification due to overfitting by the classifier, yet there is no clear performance metric to monitor which does not rely on unrealistic oracle knowledge of which points are leaking. Therefore, future research should seek out non-oracle-based performance metrics which are correlated with the ground truth fidelity of the mask. One possible strategy would be to periodically run a template attack on the highest-leakage timesteps as indicated by the AdvMask output, and early-stop training when the template attack efficacy is maximized.

Nonetheless, this technique is a promising step towards using deep learning to localize side-channel leakage with little requirement for \textit{a priori} knowledge about the nature of the leakage. Given the ever-shrinking need for human-specified implementation knowledge in side-channel attacks, it is critical to develop strategies for defending against side-channel attacks which make minimal assumptions about the specific timesteps or target variables which will be used to carry out attacks. Our technique satisfies this criterion because diverse DNN-based side-channel attacks can be adapted into classifiers and leveraged to identify which timesteps can be exploited to carry out a side-channel attack.

\section*{Acknowledgements}

We are grateful to Timur Ibrayev, Amitangshu Mukherjee, Deepak Ravikumar, and Utkarsh Saxena for helpful discussions. The authors acknowledge the support from the Purdue Center for Secure Microelectronics Ecosystem – CSME\#210205. This work was funded in part by CoCoSys JUMP 2.0 Center, supported by DARPA and SRC.

%%%% 8. BILBIOGRAPHY %%%%
\bibliographystyle{alpha}
\bibliography{abbrev3,crypto,template}
%%%% NOTES
% - Download abbrev3.bib and crypto.bib from https://cryptobib.di.ens.fr/
% - Use bilbio.bib for additional references not in the cryptobib database.
%   If possible, take them from DBLP.

\end{document}